\documentclass[journal]{IEEEtran}

\usepackage{xcolor,soul,framed} 

\colorlet{shadecolor}{yellow}
\usepackage[pdftex]{graphicx}
\graphicspath{{../pdf/}{../jpeg/}}
\DeclareGraphicsExtensions{.pdf,.jpeg,.png}

\usepackage[cmex10]{amsmath}
\usepackage{array}
\usepackage{mdwmath}
\usepackage{mdwtab}
\usepackage{eqparbox}
\usepackage{url}
\usepackage{booktabs}
\usepackage{multirow}

\begin{document}
\pagenumbering{arabic}
\bstctlcite{IEEEexample:BSTcontrol}
    \title{A Comprehensive Study of Vision Transformers in Image Classification Tasks}
    \author{Mahmoud~Khalil,
          Ahmad~Khalil,
          and~Alioune~Ngom\\
          Department of Computer Science\\
          University of Windsor,\\
          Windsor, Ontario, Canada
         }  
\maketitle

\begin{abstract}
Image Classification is a fundamental task in the field of computer
vision that frequently serves as a benchmark for gauging
advancements in Computer Vision \cite{touvron2021training}. Over the past few years, significant progress has been made in image classification due to the emergence of deep learning. However, challenges still exist, such as modeling fine-grained visual information, high computation costs, the parallelism of the model, and inconsistent evaluation protocols across datasets. In this paper, we conduct a comprehensive survey of existing papers on Vision Transformers for image classification. We first introduce the popular image classification datasets that influenced the design of models. Then, we present Vision Transformers models in chronological order, starting with early attempts at adapting attention mechanism to vision tasks followed by the adoption of vision transformers, as they have demonstrated success in capturing intricate patterns and long-range dependencies within images. Finally, we discuss open problems and shed light on opportunities for image classification to facilitate new research ideas.
\end{abstract}


%
\IEEEpeerreviewmaketitle


\section{Introduction}
\IEEEPARstart{I}{mage} classification is a fundamental task in computer vision. it has numerous real-world applications, some of which include Medical Diagnosis, Autonomous Vehicles, Surveillance, E-commerce, Agriculture, Quality Control, and Security.
Image classification involves assigning a label or category to an input image\footnote{An image is a large matrix of grayscale values, one for each pixel and color.}. It is the process of training a machine learning model to detect and classify different objects or patterns present in the image \cite{mitchell1997mcgraw}. The model takes an image as an input, and it outputs a predicted label or category based on its learned knowledge from the training dataset. 
Formally image classification task can be defined as follows:
\vspace{5pt} 
\noindent
\fbox{
    \begin{minipage}{\dimexpr\linewidth-2\fboxsep-2\fboxrule}
        \textbf{Task} \textbf{T}: Detecting and Recognizing objects\footnote{In object detection, objects are typically defined as instances of specific classes within an image.} within images \\
        \textbf{Performance measure} \textbf{P}: percent of images correctly classified \\
        \textbf{Training experience} \textbf{E}: A dataset of images containing objects with given labels
    \end{minipage}
}
\vspace{5pt} 

\begin{figure}
  \begin{center}
  \includegraphics[width=3.5in]{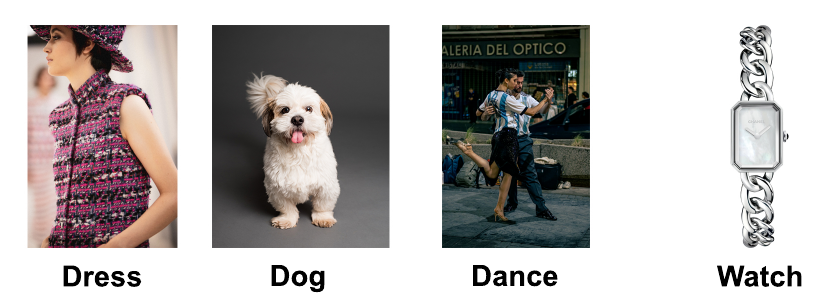}
  \caption{Visual examples of categories in popular image datasets.}\label{image_dataset_example}
  \end{center}
\end{figure}

 Fig.~\ref{image_dataset_example}, shows several images with the associated action labels, which are typical daily objects such as dress and watch.

\begin{figure}
  \begin{center}
  \includegraphics[width=3.5in]{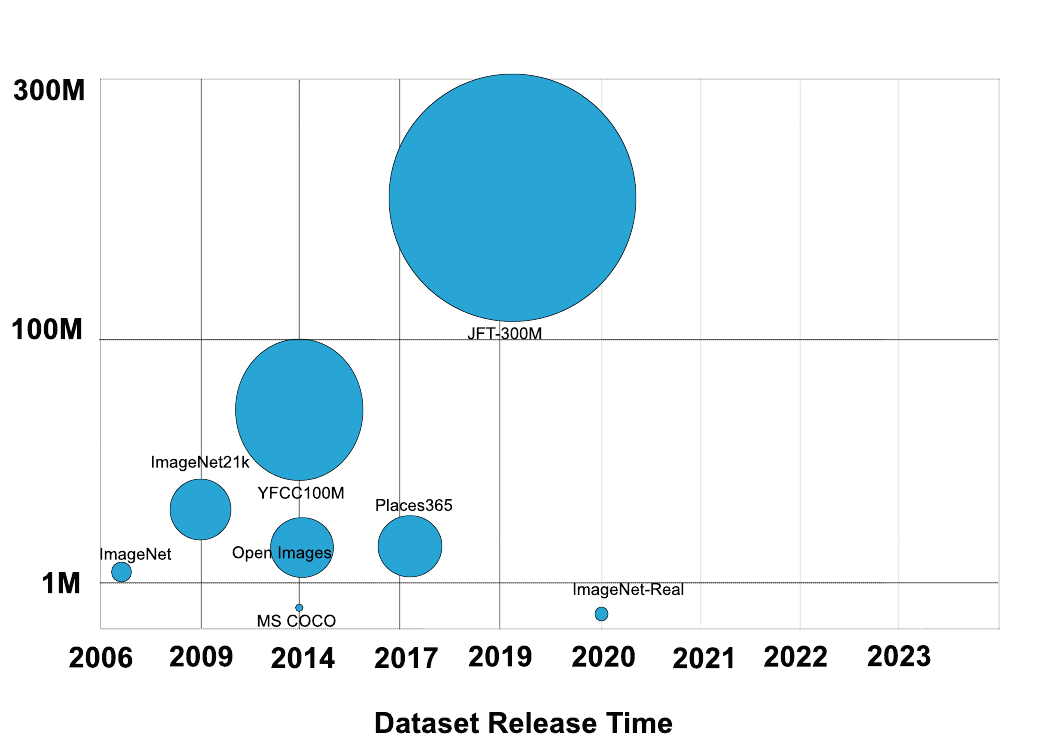}
  \caption{sizes of most popular image datasets from past 17 years. The area of a circle represents the scale of each dataset (i.e., the number of images).}\label{image_datasets_statistics}
  \end{center}
\end{figure}

Over the last decade, the success of AlexNet \cite{krizhevsky20122012} fueled a growing research interest in image classification. AlexNet's success can be attributed to its use of data augmentation and dropout regularization techniques, activation functions, and most importantly parallel computing during training \cite{chetlur2014cudnn}. Another important factor of researchers' interest in image classification is the emergence of high-quality large-scale image classification datasets \cite{krizhevsky2009learning}. We show the sizes of popular image classification datasets in Figure.~\ref{image_datasets_statistics}.

We see that both the number of images and classes increase rapidly, e.g., from 70K images over 10 classes in MNIST \cite{6296535} to 300M images over 18m classes
in JFT-300M \cite{tan2020efficientdet}. Also, the rate at which new datasets are
released is increasing: 3 datasets were released in 2014 compared to 2 released from 2006 to 2013.

Using attention as a primary mechanism for representation learning has seen widespread adoption in deep learning after \cite{vaswani2017attention}. Transformers have achieved state-of-the-art performance in a wide range of tasks \cite{devlin2018bert, brown2020language, liu2019roberta}. Recently, they have also been applied to several computer vision tasks including image classification \cite{Bertasius2021IsSA, akbari2021vatt}. 

There are several Transformer models that have been used in image classification tasks, including:

\begin{itemize}
  \item 
  Vision Transformer (ViT): ViT \cite{dosovitskiy2020image} is a transformer-based image classification model that was introduced in 2020 by Google. It uses a standard transformer encoder to process image patches extracted from the input image.
  \item 
  Swin Transformer: Swin \cite{liu2021swin} Transformer is a hierarchical transformer-based model that can process images of any size. It divides the input image into non-overlapping patches, which are then processed in a hierarchical manner by multiple transformer layers.
  \item 
  DeiT: DeiT \cite{touvron2021training}  stands for "Dense Encoder-Decoder Transformers." It is a transformer-based model that uses a combination of encoder and decoder layers to process image patches.
  \item 
  CaiT: CaiT \cite{touvron2021going} stands for "Cross-Attention Image Transformer." It is a transformer-based model that uses cross-attention to combine information from different patches of the input image.
  
  \item 
  iGPT: The iGPT \cite{chen2020generative} model is a type of Generative Pre-trained Transformer that was introduced in a research paper by the OpenAI team in 2021. "iGPT" stands for "image GPT", which means that the model has been trained to generate images in addition to text. Unlike previous GPT models, which were primarily designed for generating natural language text, iGPT has been trained to generate high-resolution images from textual descriptions.
\end{itemize}

\begin{figure*}
  \begin{center}
  \includegraphics[width=7in]{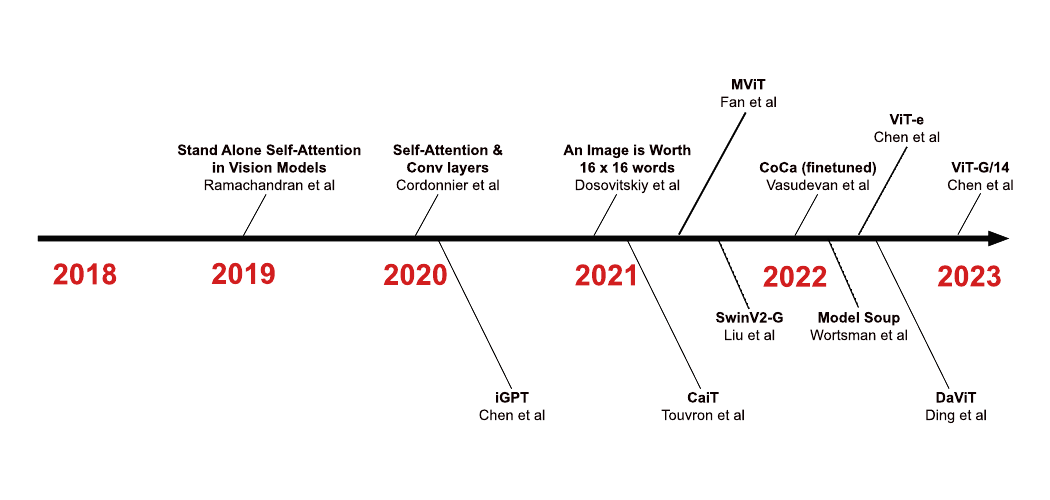}
  \caption{A chronological overview of recent representative work in image classification .}\label{image_classification_chronolgical}
  \end{center}
\end{figure*}

 In Fig.~\ref{image_classification_chronolgical}, we present a chronological overview of recent representative work

The Stand-Alone Self-Attention in Vision Models \cite{ramachandran2019stand} marks an early departure from convolutional approaches in computer vision. Three notable trends emerged subsequently. The first trend, initiated by a pivotal paper demonstrating the viability of attention layers as substitutes for convolution \cite{cordonnier2019relationship}, involves extracting $2 \times 2$ patches from the input image and applying full self-attention. Following the success of applying transformers directly to images, exemplified by the Vision Transformer (ViT) \cite{dosovitskiy2020image}, various subsequent models like CaiT \cite{touvron2021training}, Swin \cite{liu2021swin}, Coca \cite{yu2022coca}, among others, have been developed.

The second trend involves adopting a multimodal approach to facilitate generalization to novel categories without the need for retraining. Notable examples include iGPT \cite{chen2020generative}, CLIP \cite{radford2021learning}, Model Soup \cite{wortsman2022model}, among others.

Finally, the third trend focused on computational efficiency to increase the throughput
of existing ViT models without needing to train. Examples include ToMe \cite{bolya2022token}, Hiera \cite{ryali2023hiera}. 

Despite there being many papers using vision transformers for image classification,\cite{dosovitskiy2020image,liu2021swin,touvron2021going}, convolutional neural networks \cite{krizhevsky2017imagenet,he2016deep,szegedy2015going}, particularly ResNet50 \cite{he2016deep}, dominated the image classification literature, due to their high accuracy and good robustness. Nevertheless, convolutional neural networks incur substantial computational expenses \cite{raj2016towards,lee2017parallel,pourghassemi2020brief}. Additionally, parallelizing a CNN model is challenging due to the inherently tightly coupled structure of CNNs \cite{wang2023model}.

The dominant approach to creating convolutional neural network systems  \cite{peng2022survey} is to collect a dataset of training examples demonstrating correct behavior for a desired task, train a system to imitate these behaviors, and then test its performance on independent and identically distributed (IID) held-out examples. This has served well to make progress on narrow experts rather than competent generalists.

First, from a practical perspective, the need for a large dataset of labeled examples for every new task limits the applicability of models.

Second, A major limitation of convolutional neural networks is if we want to classify other types of images or objects that are not part of the training distribution, we would need to train a new convolutional neural network on a dataset that is relevant to the specific types of images we want to classify.
Alternatively, we can use transfer learning, which involves fine-tuning a pre-trained CNN on a new dataset. 

On the other hand, Transformer models are input agnostic, task agnostic, and architecture agnostic\footnote{Transformers are highly versatile and can be applied to a wide range of tasks}, and architecture agnostic\footnote{The fundamental principles and techniques used in transformer-based models can be applied across various transformer architectures.}. The Transformer architecture is built based on parallelization, allowing for efficient processing and scalability \cite{vaswani2017attention}. 
These transformer-based models have shown promising results on several image classification benchmarks and are being actively researched to improve their performance further.

Despite the large number of attention-based models for image classification, there is no comprehensive survey dedicated to these models. Previous survey papers either put more effort into convolutional neural networks \cite{9730565, plested2022deep, he2020deep, rawat2017deep} or focus on broader topics such as video processing \cite{han2022survey},
medical imaging \cite{shamshad2023transformers}, visual learning understanding \cite{yang2022transformers} and
object detection, action recognition,  segmentation \cite{khan2022transformers}. In this paper:

\begin{itemize}
\item 
 We review papers on attention mechanisms for image classification. We walk the readers through the recent advancements chronologically and systematically.
\item 
We comprehensively review papers on Vision Transformers for image classification. We walk the readers through the recent advancements chronologically and systematically, with popular papers explained in detail.
\item 
We elaborate on challenges, open problems, and opportunities in this field to facilitate future research.

\item
We benchmark widely adopted methods on the same set of datasets in terms of both accuracy and efficiency.
\end{itemize}

The survey is organized as follows. We first describe popular datasets used for benchmarking and existing challenges in section 2. Then we walk the reader through the background of the attention mechanism for vision in section 3. Then we present recent advancements using transformers for image classification in section 4, which is the major contribution of this survey. In section 5, we evaluate widely adopted approaches
on standard benchmark datasets, and provide discussions and future research opportunities in section 6.

\section{Datasets and Challenges}
\subsection{Datasets}
The most current theory of machine learning rest on the crucial assumption that the distribution of the training dataset is identical to the distribution of the future data over which the final system performance $\mathbf{P}$ must be measured \cite{mitchell2007machine}. Transformers lack some of the inductive biases inherent to CNNs \cite{dosovitskiy2020image}, such as translation equivariance and locality, and therefore do not generalize well when trained on insufficient amounts of data, which means we need large-scale annotated datasets to learn effective models.

Image classification datasets are built by collecting large numbers of images and manually labeling them with corresponding class labels. The process involves several steps:

(1) Image Collection: The first step is to collect a large number of images relevant to the classification task. These images can be obtained from various sources such as the internet, public databases, or by capturing images using cameras or other imaging devices.

(2) Image Labeling: The collected images are then manually labeled with class labels by experts or using crowd-sourcing methods. Each image is assigned a label that corresponds to the object or category that the image represents.

(3) Data Cleaning: The labeled images are then cleaned by removing duplicates, irrelevant images, or images with low quality or resolution.

(4) Data Splitting: The dataset is split into three sets - training, validation, and testing sets. The training set is used to train the model, while the validation set is used to tune the hyperparameters of the model. The testing set is used to evaluate the performance of the model on unseen data.

(5) Data Augmentation: Data augmentation techniques such as image cropping, flipping, rotation, and color shifting are applied to increase the diversity of the dataset and improve the generalization performance of the model.

(6) Preprocessing: The images are preprocessed by resizing, normalizing, and converting them to a standardized format suitable for input to the machine learning model.
Below, we review the most popular large-scale image classification datasets in Table~\ref{image_classification_datasets_table} and Figure~\ref{image_datasets_statistics}.

\begin{table}[!h]
\begin{center}
\begin{tabular}{ |p{3cm}||p{1cm}|p{1cm}|p{1cm}|p{1cm}| }
 \hline
 \multicolumn{5}{|c|}{Dataset List} \\
 \hline
 Dataset & Year & \#Samples & \#Classes & Access \\
 \hline
 ImageNet \cite{deng2009imagenet, russakovsky2015imagenet}& 2006& 1.2m& 1k & Public\\
 VOC07\cite{pascal-voc-2007}& 2007& 9k& 20 & Public\\
 Oxford Flowers \cite{nilsback2008automated} & 2008 & 8k & 102 & Public \\ 
 CIFAR-10 \cite{krizhevsky2009learning} & 2009 & 60k & 10 & Public \\ 
 CIFAR-100 \cite{krizhevsky2009learning} & 2009 & 60k & 100 & Public \\ 
 ImageNet-21k \cite{ridnik2021imagenet}& 2009& 14m& 21k & Public\\
 Oxford-IIIT Pets \cite{parkhi2012cats} & 2012 & 7k & 37 & Public \\
 YFCC100M\cite{thomee2016yfcc100m}& 2014& 99.2m& 21k & Public\\
 MS COCO\cite{lin2014microsoft}& 2014& 330k& 80 & Public\\
 Open Images\cite{thomee2016yfcc100m}& 2014& 9m& 21k & Public\\
 Places365\cite{zhou2017places}& 2017& 10m& 365 & Public\\
 iNat18 \cite{van2018inaturalist}& 2018& 859k& 5k & Public\\
 JFT-300M\cite{tan2020efficientdet}& 2019& 300m& 18m & Private\\
 ImageNet-ReaL\cite{karpathy2014large}& 2020& 300k& 314 & Public\\ 
 \hline
\end{tabular}
\caption{\label{image_classification_datasets_table}A list of popular datasets for image classifications.}
\end{center}
\end{table}

\subsubsection{ImageNet}
ImageNet \cite{deng2009imagenet, russakovsky2015imagenet} is a large-scale image database created by researchers at Princeton University, initially designed for the purpose of object recognition research in computer vision. It was first introduced in 2009 and consists of over 1.2 million images that are labeled with one of 1,000 object categories. Each image in the dataset has been manually annotated with a bounding box that indicates the location of the object within the image.
ImageNet has become one of the most widely used image datasets in the field of computer vision, and it has been used to train and evaluate a wide range of deep learning models for image classification, object detection, and other related tasks. The dataset was created with the goal of advancing the state-of-the-art in image recognition, and it has been used as the basis for the annual ImageNet Large Scale Visual Recognition Challenge (ILSVRC), which is a competition that evaluates the performance of different image recognition algorithms on a set of standardized tasks using the ImageNet dataset.
Since its creation, ImageNet has been expanded and extended to include more images and categories, such as the ImageNet-21k dataset, which contains over 14 million images across more than 21,000 categories, and the ImageNet-R dataset, which is a subset of ImageNet that has been specifically designed to be more robust to changes in image resolution, lighting, and other factors.

\subsubsection{ImageNet-21k}
ImageNet-21k \cite{ridnik2021imagenet} is an extension of the original ImageNet dataset, which contains over 14 million images across more than 21,000 categories. The original ImageNet dataset was created in 2009 and has been widely used as a benchmark for image classification algorithms.
In 2017, researchers at Stanford University released ImageNet-21k, with the goal of increasing the diversity of images and categories included in the dataset. ImageNet-21k includes more fine-grained categories, such as different breeds of dogs or types of flowers, as well as categories that were not included in the original dataset, such as different types of vehicles and tools.
ImageNet-21k is significantly larger and more diverse than the original ImageNet dataset and presents a more challenging task for image classification algorithms. It has been used to train and evaluate the performance of deep neural networks on a wider range of image classification tasks. However, due to its large size and complexity, it also requires significant computing resources and time to train models on this dataset.

\subsubsection{YFCC100M}
YFCC100M \cite{thomee2016yfcc100m} (Yahoo Flickr Creative Commons 100 Million) is a large-scale dataset of images and videos that were created by Yahoo in collaboration with Flickr. It was released in 2014 and is one of the largest publicly available multimedia datasets.
The YFCC100M dataset contains over 99.2 million images and 0.8 million videos, with a total size of over 1 TB. The images and videos cover a wide range of categories, including people, animals, objects, scenes, and more, and they were collected from a diverse set of sources, including amateur photographers and professional news organizations.
One of the unique features of YFCC100M is that it includes rich metadata for each image and video, including information about the camera used, the location and time the media was captured, and other contextual information. This makes it a valuable resource for research in computer vision and machine learning, as well as for applications such as image search and recommendation systems.
YFCC100M has been used for a variety of research applications, including image classification, object detection, and image retrieval. It is freely available for research purposes, and it continues to be updated and expanded over time.

\subsubsection{MS COCO}
MS COCO \cite{lin2014microsoft} (Microsoft Common Objects in Context) is a large-scale image dataset created by Microsoft Research in collaboration with several other organizations. It was introduced in 2014 and has become a widely used benchmark for image recognition, object detection, and segmentation.
The MS COCO dataset contains over 330,000 images, each annotated with object-level labels, object detection bounding boxes, and segmentation masks for common objects in context. The images cover a wide range of categories, including people, animals, objects, and scenes, and are designed to be more challenging than other popular image datasets, such as ImageNet.
MS COCO has been used as a benchmark for a variety of computer vision tasks, including object detection, image segmentation, and image captioning. It has also been used to train and evaluate a wide range of deep learning models, and it has spurred significant progress in the field of computer vision research.
One of the unique features of MS COCO is that it includes annotations for a wide range of object sizes and occlusion levels, making it a valuable resource for developing more robust and accurate object detection and segmentation models. The dataset is freely available for research purposes, and it continues to be updated and expanded over time.

\subsubsection{Open Images}
Open Images \cite{thomee2016yfcc100m} is a large-scale dataset of annotated images that was created by Google in collaboration with several other organizations. It was introduced in 2016 and has been widely used for research in computer vision and machine learning.
The Open Images dataset contains over 9 million images, each annotated with object-level labels, object detection bounding boxes, and visual relationships between objects. The images in the dataset cover a wide range of categories, including people, animals, objects, and scenes.
One of the unique features of Open Images is that it includes a large number of images with multiple objects, which makes it useful for training and evaluating object detection models. The dataset also includes a set of challenge tasks, such as object detection and visual relationship detection, which have been used to evaluate the performance of different computer vision algorithms.
Open Images has been used as a benchmark for a wide range of computer vision tasks, and it has been used to train and evaluate a variety of deep learning models, including those for object detection, image segmentation, and image classification. The dataset is freely available for research purposes, and it continues to be updated and expanded over time.

\subsubsection{Places365}
Places365 \cite{zhou2017places} is a large-scale image dataset of scenes and places that was created by researchers at the MIT Computer Science and Artificial Intelligence Laboratory (CSAIL). It was introduced in 2017 and has become a widely used benchmark for scene recognition and image classification.
The Places365 dataset contains over 1.8 million images, each labeled with one of 365 scene categories. The images cover a wide range of indoor and outdoor scenes, including landscapes, buildings, streets, and interiors, and are designed to be more challenging than other popular scene datasets, such as SUN397 and Places205.
One of the unique features of Places365 is that it includes multiple images of the same scene, captured under different lighting and weather conditions. This makes it a valuable resource for developing more robust and accurate scene recognition models.
Places365 has been used as a benchmark for a variety of computer vision tasks, including scene recognition, image classification, and transfer learning. It has also been used to train and evaluate a wide range of deep learning models, and it has spurred significant progress in the field of computer vision research.
The dataset is freely available for research purposes, and it continues to be updated and expanded over time.

\subsubsection{JFT-300M}
JFT-300M \cite{tan2020efficientdet} (The JFT-300M dataset) stands for the "Google's JFT-300M" dataset, which is a large-scale image dataset containing approximately 300 million images. This dataset was created by Google as a part of their research in computer vision and machine learning, and it was made publicly available in 2019.
JFT-300M is one of the largest image datasets currently available, and it contains images across a wide range of categories, including people, animals, objects, scenes, and more. However, unlike other popular image datasets such as ImageNet, JFT-300M is not annotated with object categories. Instead, it is designed to be used for pre-training deep neural networks for various computer vision tasks, such as object detection, image segmentation, and image classification.
The JFT-300M dataset is challenging due to its large size and the diversity of images and categories, making it a valuable resource for advancing research in the field of computer vision. However, due to its massive size, it requires significant computing resources and time to use effectively.

\subsubsection{ImageNet-ReaL}
ImageNet-ReaL \cite{karpathy2014large} (ImageNet-Real) is a dataset that was introduced in 2020 as an extension of the original ImageNet dataset, with the goal of better reflecting the challenges and complexities of real-world image classification tasks. The "ReaL" in ImageNet-ReaL stands for "Realistic and Large-scale" image classification.
Unlike the original ImageNet dataset, which primarily consists of high-quality images captured under controlled conditions, ImageNet-ReaL includes images that are more representative of the types of images encountered in real-world applications. The images in ImageNet-ReaL are taken from a wide range of sources, including social media, news articles, and online shopping websites, and are often low-quality, noisy, or occluded.
ImageNet-ReaL contains over 300,000 images across 314 categories, and it has been used as a benchmark for evaluating the performance of deep learning models on real-world image classification tasks. The dataset has been found to be significantly more challenging than the original ImageNet dataset, and it requires models to be more robust to variations in image quality, lighting, and other factors that are common in real-world images.

\subsection{Challenges}
Developing effective image classification algorithms comes with several challenges. Here are some of the major challenges faced in this field:

\begin{itemize}
    \item 
    Variability in visual appearance: Images can exhibit substantial variations in lighting conditions, viewpoint, scale, occlusions, and background clutter. These factors make it challenging to accurately classify objects or scenes, as the same object can appear differently in different images \cite{lenc2015understanding, mahendran2015understanding, sung2018learning}.
    \item
    Large-scale datasets and computational requirements: Image classification often requires large-scale labeled datasets for training deep learning models effectively. Acquiring and annotating such datasets can be time-consuming and expensive. Moreover, training deep neural networks on these large datasets can be computationally intensive and may require specialized hardware \cite{sejnowski2020unreasonable, szegedy2016rethinking}.
    \item
    Overfitting and generalization: Deep learning models are prone to overfitting, where they memorize the training data without effectively generalizing to new, unseen images. Achieving good generalization is crucial to ensure that the model can accurately classify images from different sources or distributions \cite{zhang2021understanding, wan2013regularization, keskar2017improving, he2019bag}.
    \item
    Limited labeled data: In some cases, labeled training data may be scarce or difficult to obtain. Labeling large datasets can be a labor-intensive process, and certain image categories may require domain expertise or subjective judgments. Limited labeled data can hinder the development of accurate classifiers \cite{odena2016semi, liu2018leveraging}.
    \item
    Fine-grained classification: Traditional image classification algorithms focus on discriminating broad categories like dogs vs. cats. However, distinguishing between fine-grained categories within a class (e.g., different bird species) is more challenging due to subtle visual differences. Capturing these fine-grained distinctions requires more nuanced models and specialized techniques \cite{kumar2019bird, lin2015bilinear}.
    \item
    Class imbalance: In real-world datasets, certain classes may have significantly more samples than others, leading to class imbalance. Class imbalance can bias the model towards the majority class and affect its ability to correctly classify minority classes, which are often the ones of particular interest \cite{he2009learning, han2019deep}.
    \item 
    Robustness to adversarial examples: Adversarial examples are carefully crafted inputs that are designed to mislead the classifier, even though they may appear similar to humans. Building image classifiers that are robust to such adversarial attacks is an ongoing challenge \cite{huang2017adversarial}.

\end{itemize}

Addressing these challenges often involves a combination of algorithmic advancements, larger and more diverse datasets, architectural improvements in deep learning models, regularization techniques, data augmentation, transfer learning, and other strategies to improve the accuracy and robustness of image classification algorithms.
\section{background on attention mechanism for vision}
Machine learning model architectures based on neural networks have been very successful in improving the state of the art on complex image classification benchmark datasets \cite{lyu2019advances, sermanet2013overfeat}. However, these architectures require significant computational resources for both training and testing, despite achieving high recognition accuracy. The current state-of-the-art convolutional neural networks in use can take several days to train on multiple GPUs, even when input images are downsampled to reduce computation \cite{lyu2019advances}. The primary computational cost for these models arises from convolving filter maps with the entire input image. This means that the computational complexity of these models is at least linear to the number of pixels in the input image.
One important property of human perception is that one does not tend to process a whole scene in its entirety at once. Instead, humans focus attention selectively on parts of the visual space to acquire information when and where it is needed and combine information from different fixations over time to build up an internal representation of the scene\cite{rensink2000dynamic}.

Concentrating the computational power on specific parts of a scene helps to reduce the amount of data that needs to be processed, as fewer pixels are involved. Additionally, it simplifies the task at hand because the object of interest can be placed at the center of focus, while irrelevant features of the visual environment, which are referred to as "clutter," can be disregarded since they lie outside the fixated region.

One of the early attempts to mimic human attention is saliency detectors (e.g. \cite{itti1998model}). Saliency detectors indeed capture some of the properties of human eye movements, but they typically do not integrate information across fixations, their saliency computations are mostly hardwired, and they are based on low-level image properties only, usually ignoring other factors such as semantic content of a scene and task demands \cite{torralba2006contextual}.

Modern attention research, however, is guided by many mathematical and engineering disciplines, and the goal of the attention mechanism is not to model human attention. It is best to think of the attention mechanism as a function approximation of a database select query \cite{bahuleyan2017variational}, rather than as a model of the human brain's attention function.

\begin{figure}
  \begin{center}
  \includegraphics[width=3in]{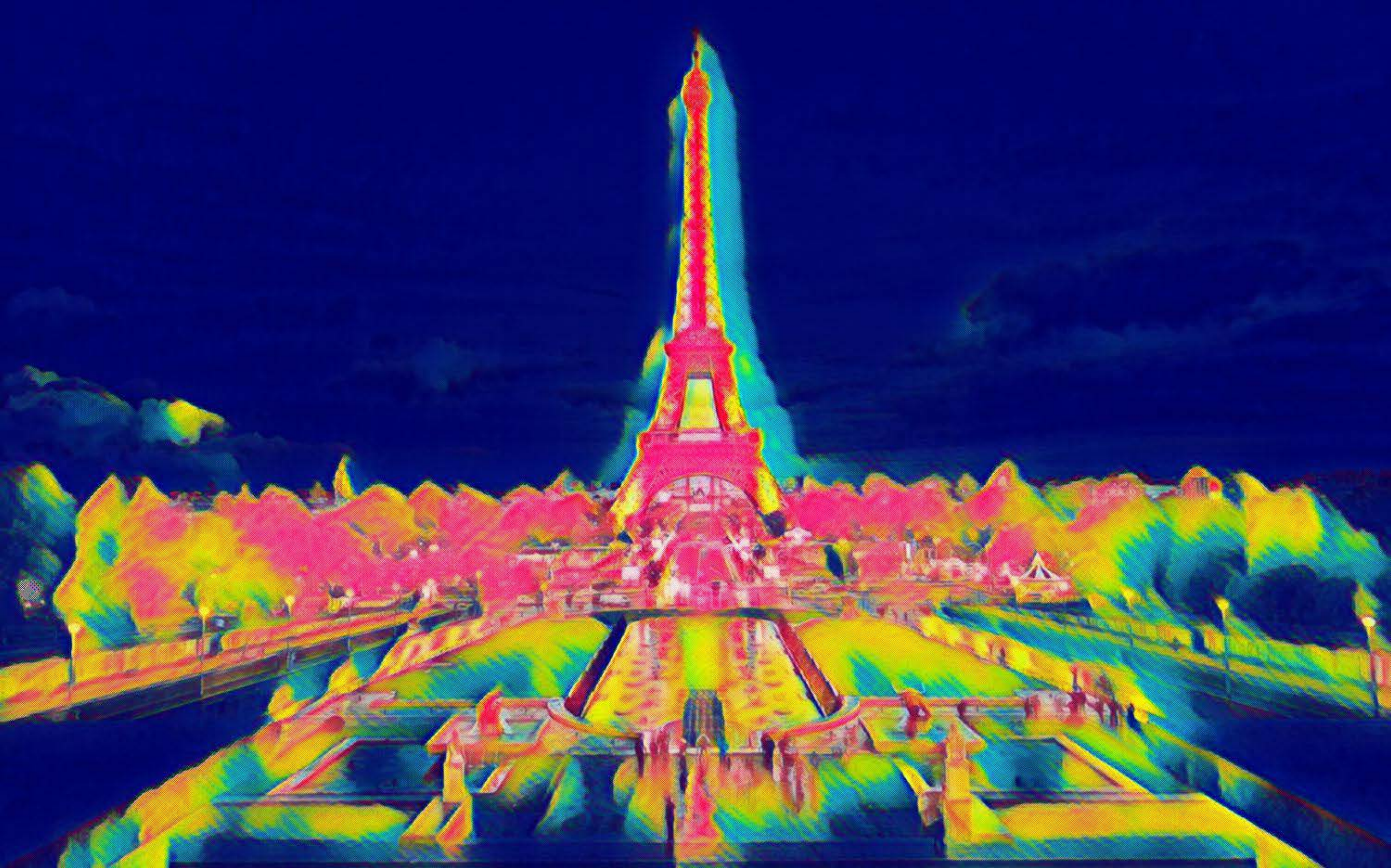}\\
  \caption{Attention to highlight parts that contribute to classification. The heat map corresponds to the weights of the attention mechanism.}\label{tour_eiffel}
  \end{center}
\end{figure}

The concept of attention was first studied in the context of computer vision, as a way of highlighting important parts of an image that contribute to the desired outcome \cite{mnih2014recurrent}. For example, in image classification, if we trained some model to do image classification, an interesting question is when the model outputs a class, how can we trust that the model outputs the correct class? Here we can use attention to essentially see which pixels are aligned with the concept of the class. In Fig.~\ref{tour_eiffel}, attention mechanisms show which pixels are aligned with the concept of the Eiffel Tower.

\begin{figure}
  \begin{center}
  \includegraphics[width=3in]{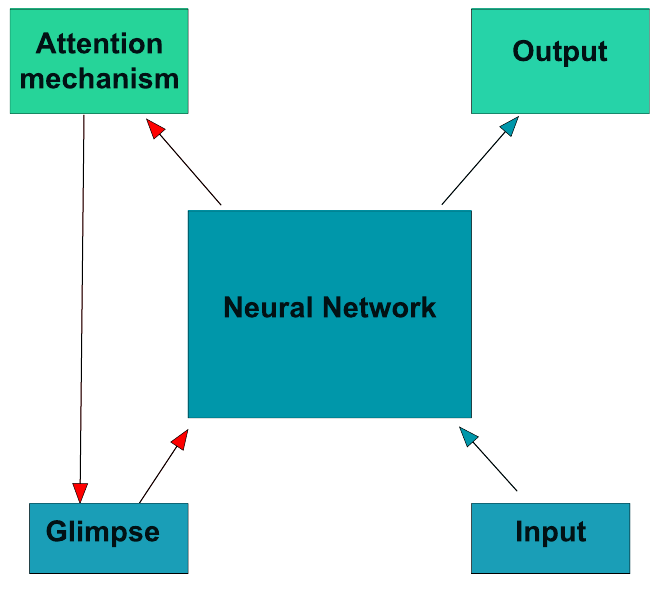}\\
  \caption{Neural attention models.}\label{neural_attention_models}
  \end{center}
\end{figure}

There are many kinds of neural attention models, and they all have kind of similar formats. The network receives input and produces output as usual. The attention model creates a glimpse that is passed to the network as an extra input in the next time step. The entire system is recurrent even if the network is not. Fig.~\ref{neural_attention_models}, shows the general format of neural attention models.

\section{Types of Attention Mechanisms}
\subsection{Stochastic “Hard” Attention}
\begin{figure}
  \begin{center}
  \includegraphics[width=3in]{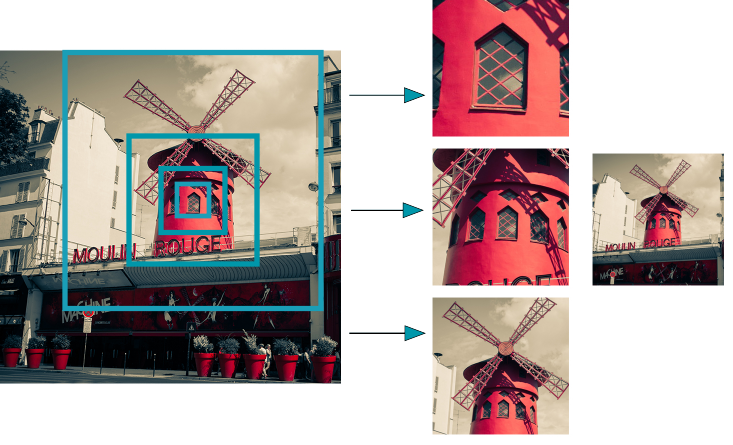}\\
  \caption{Glimpse Sensor: Given the coordinates of the glimpse and an input image, the sensor extracts a retina-like representation centered certain position in the image that contains multiple resolution patches.}\label{hard_attention_sensor}
  \end{center}
\end{figure}

  is a discrete form of attention that selects a single input feature or region at each time step, based on the learned attention distribution \cite{butko2009optimal}. Hard Attention is a sequential decision process of a goal-directed agent. At each point in time, the agent observes the environment only via a bandwidth-limited sensor see Fig.~\ref{hard_attention_sensor}, i.e. it never senses the environment in full. The agent can, however, actively control how to deploy its sensor resources (e.g. choose the sensor location). While Hard Attention-based models are non-differentiable (cannot be trained using gradient), they can be trained using reinforcement learning methods to learn task-specific policies. At each step, the agent receives a scalar reward (which depends on the actions the agent has executed and can be delayed), and the goal of the agent is to maximize the total sum of such rewards \cite{mnih2014recurrent}.
  To understand hard attention, let's consider an example of an image classification model. 
  The Hard attention model in this model is built around recurrent neural networks. At each timestep, it analyzes the sensor data, incorporates information over time, and makes decisions on actions and sensor deployment for the next time step.
  \textbf{Glimpse Sensor:}
  At each time step t, the agent is provided with a (partial) observation of the environment represented as an image $x_t$. However, the agent's access to this image is restricted, and it can only extract information from $x_t$ using its sensor $\mathcal{P}$, which has limited bandwidth. For instance, the agent can choose to focus the sensor on specific regions or frequency bands of interest \cite{larochelle2010learning}.

\textbf{Internal state:}
The agent possesses an internal state that captures a summarized representation of information derived from past observations. This internal state encompasses the agent's understanding of the environment and plays a crucial role in determining the appropriate actions and sensor deployment. The recurrent neural network generates this internal state using its hidden units h\_t, which are continually updated by the core network over time:
\[h_t = f_h (h_{t-1} , g_t; \theta_h)\]
The glimpse feature vector $g_t$ serves as the external input to the network.

\textbf{Actions:}
At each step, the agent undertakes two actions. Firstly, it determines how to deploy its sensor through the sensor control $l_t$, and secondly, it performs an action within the environment that may impact the state of the environment. The nature of the environmental action varies depending on the specific task at hand. In this particular example, the location actions are probabilistically selected from a distribution, which is parameterized by the location network $f_l(h_t ; \theta_l)$ at time t: \[l_t \sim p (.|f_l(h_t;\theta_l))\]. Similarly, the environment action $a_t$ is drawn from a distribution conditioned on the output of a second network, at 
\[a_t \sim p(.|f_a(h_t; \theta_a))\]
For classification tasks, a softmax output formulation is employed, while in dynamic environments, the exact formulation relies on the action set specific to that particular environment (e.g., joystick movements, motor control, etc.). Additionally, our model can be enhanced with an extra action that determines when the agent should cease taking glimpses. This feature can be utilized, for instance, to train a cost-sensitive classifier by assigning a negative reward to each glimpse, thereby encouraging the agent to balance correct classifications with the cost of taking additional glimpses.

\textbf{Reward:}
Once the agent performs an action, it obtains a fresh visual observation of the environment $x_{t+1}$ and receives a reward signal $r_{r+1}$. The objective of the agent is to maximize the cumulative sum of the reward signal, denoted as 
$R = \sum_{t = 1}^{T} r_t$. This reward signal is typically sparse and delayed, meaning that significant rewards are infrequent and might be received only after a certain number of steps. For instance, in the context of object recognition, $r_T = 1$ if the object is correctly classified after T steps, and 0 otherwise.

\subsection{Deterministic “Soft” Attention}
  Soft attention is a function approximation that mimics the retrieval of a value $v_i$ for a query $q$ based on a key $k_i$ in the database in a probabilistic and fuzzy way using the following template equation
  \[attention(q,k,v) = \sum_{i}^{} similarity(q,k_i) \times v_i\]

  There are many kinds of variants of Soft Attention but they all come back to the above basic template.
  Soft Attention is a continuous distribution over the input features or regions and is learned from data by the neural network during training \cite{bahdanau2014neural}. The output of the attention mechanism is a weighted sum of the input features or regions, where the weights are determined by the learned attention distribution.

\subsection{Self-Attention (Scaled Dot Product Attention)}
This type of soft attention mechanism is used within the same input sequence or feature map. Self-attention is commonly used in NLP tasks, such as machine translation and text classification, where the network needs to selectively attend to different parts of the input sequence to generate an output.

    Mathematically, self-attention can be expressed as follows:
    
    Let $x$ be the input sequence of length $T$, where each element $x_t$ is a feature vector of dimension $d$. 

\[Attention(Q,K,V) = softmax(\frac{QK^T}{\sqrt{d_k}})V \]

where $Q$ and $K$ are learned query and key matrices, respectively, of dimension $d_k \times n$. The softmax function ensures that the weights sum to one, making them a probability distribution over the input sequence. The division by $\sqrt{d_k}$ is a scaling factor that stabilizes the gradients during training.

\begin{figure}
  \begin{center}
  \includegraphics[width=3in]{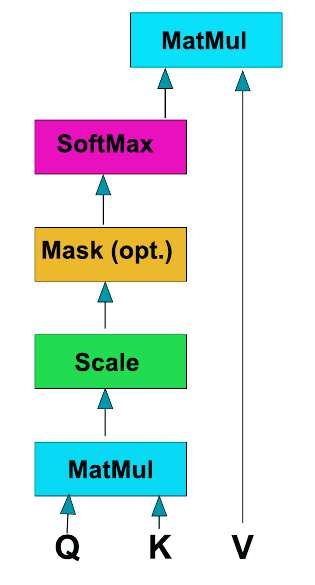}\\
  \caption{Self Attention (Scaled Dot Product Attention)}\label{self_attention}
  \end{center}
\end{figure}

This process is repeated for each time step t in the sequence, creating a set of updated hidden states h' that capture the dependencies between different elements of the input sequence. These hidden states can then be used for downstream tasks, such as sequence classification or generation. Self-attention has been shown to be effective in a wide range of natural language processing tasks, including machine translation, language modeling, and question-answering.
Fig.~\ref{self_attention}, shows the format of the self-attention layer. 

\subsection{Multi-Head Attention}
Multi-headed attention is an extension of the Self Attention mechanism. Multi-headed attention enables the model to attend to different parts of the input sequence simultaneously and learn multiple representations of the input at different levels of granularity.

In multi-headed attention, the original input is transformed into multiple query, key, and value vectors sets, where each set is referred to as a "head." The transformation is typically achieved by linear projections, which map the input to different subspaces. These projected vectors serve as inputs for separate attention mechanisms or attention heads.

Each attention head computes attention weights and generates a context vector independently. This means that each head focuses on different aspects or relationships within the input sequence. The attention weights for each head are computed through a similar process as in the standard attention mechanism, using the query, key, and value vectors specific to that head.

After the attention weights are computed for each head, the context vectors from all the heads are concatenated or linearly combined to create a combined representation. This combined representation is then passed through another linear projection to produce the final output.

The benefit of multi-headed attention is that it allows the model to capture different types of information or relationships within the input sequence. Each head can specialize in attending to different aspects, such as word order, semantic meaning, or syntactic structure. By attending to multiple perspectives simultaneously, the model gains a richer understanding of the input and can better capture complex patterns and dependencies.

Multi-headed attention has been a key component in transformer models, such as the Transformer architecture, which have achieved state-of-the-art performance in various NLP tasks, including machine translation, text generation, and language understanding.

\subsection{Global Attention}
Global attention is a mechanism used in artificial neural networks, particularly in the field of natural language processing (NLP), to capture and incorporate information from all parts of a sequence into the learning process. It is often applied in tasks such as machine translation, text summarization, and question answering.

In NLP, sequences are typically represented as a series of vectors, where each vector represents a word or a token in the sequence. Global attention allows the model to focus on different parts of the input sequence during the encoding or decoding process by assigning weights or importance scores to each vector.

The global attention mechanism works by computing a compatibility score between a "query" vector and each vector in the sequence. The query vector is typically derived from the hidden state of the recurrent neural network or a decoder in the model. The compatibility scores indicate how relevant each vector is to the query.

To determine the attention weights, these compatibility scores are transformed into attention weights using a softmax function. The attention weights determine the importance of each vector in the sequence relative to the query. Higher weights imply that the corresponding vectors are more relevant and should receive more attention.

Once the attention weights are computed, a weighted sum of the sequence vectors is taken to produce a context vector, which represents the attended information. This context vector is then used in further computations, such as generating the next word in a sequence or providing additional information to downstream tasks.

Global attention allows the model to dynamically focus on different parts of the input sequence depending on the context and the specific task at hand. By attending to the most relevant parts of the sequence, it helps the model improve its performance in understanding and generating language.

\subsection{Local Attention}
Local attention is a variant of the attention mechanism used in artificial neural networks, specifically in the field of natural language processing (NLP). Unlike global attention, which considers all parts of a sequence when computing attention weights, local attention focuses only on a limited or localized region of the sequence.

In local attention, instead of computing attention weights for every vector in the sequence, attention is restricted to a subset or window of vectors around a particular position. This window can be fixed or dynamic, depending on the specific implementation.

The idea behind local attention is to reduce the computational complexity associated with global attention, particularly when dealing with long sequences. By limiting the attention to a local region, the model can prioritize relevant information while ignoring less relevant or distant parts of the sequence.

The process of computing attention weights in local attention is similar to global attention. A query vector, typically derived from the hidden state of the neural network, is used to compute compatibility scores with the vectors within the window. The scores are then transformed into attention weights using a softmax function, assigning higher weights to more relevant vectors within the window.

The context vector is then computed as a weighted sum of the vectors within the window, using the attention weights. This context vector represents the attended information, which is utilized for further computations in the model.

Local attention is particularly useful when dealing with long sequences, such as lengthy documents or paragraphs, where considering the entire sequence may be computationally expensive or unnecessary. By restricting attention to local regions, models can still capture relevant information while achieving computational efficiency.

There are many other variations and combinations of attention mechanisms that have been proposed in the literature, and the choice of attention mechanism often depends on the specific task and application at hand.

\begin{figure*}
  \begin{center}
  \includegraphics[width=7in]{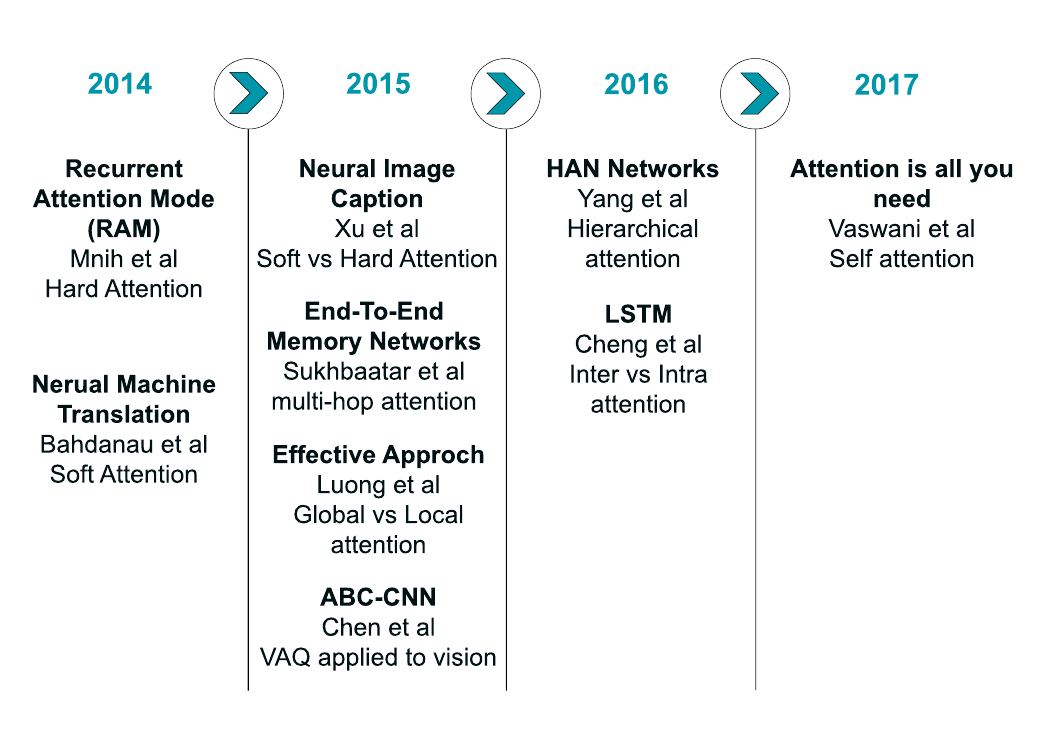}
  \caption{A chronological overview of attention mechanism.}\label{attention_mechanism_chronolgical}
  \end{center}
\end{figure*}

\section{Using Deep Learning for Image Classification}

In this section, we review deep learning-based methods
for image classification from 2014 to the present, introduce the related earlier work in context.

\subsection {From Convolutional Neural Networks to Transformers}
 Recently, a new architecture, known as Transformers, has shown the potential to take over CNNs as the primary architecture for computer vision tasks. In this section, we will explore how computer vision is transitioning from CNNs to Transformers models.

\begin{table}[ht]
\centering
\begin{tabular}{|p{1cm}|p{2.5cm}|c|c|c|}
\hline
\textbf{Model} & \textbf{Characteristics} & \textbf{Params (M)} & \textbf{FLOPs (G)} & \textbf{Year} \\ \hline
VIT & Strong attention mechanism & 86 & 5.6 & 2020 \\ \hline
SWIN & Hierarchical transformer architecture & 94 & 6.8 & 2021 \\ \hline
MVIT & Multiple vision transformers & 92 & 6.5 & 2021 \\ \hline
DEIT & Token-based training, no CNN & 86 & 5.2 & 2020 \\ \hline
CAIT & Class-agnostic instance training & 88 & 5.9 & 2021 \\ \hline
HERA & Heterogeneous transformer modules & 96 & 7.2 & 2022 \\ \hline
\end{tabular}
\caption{Summary of Different Vision Transformer Models Based on Characteristics.}
\label{tab:vision_transformers}
\end{table}

 Transformers were initially introduced in the context of natural language processing (NLP) \cite{vaswani2017attention} to model the relationships between words in a sentence. Unlike CNNs, Transformers use an attention mechanism that allows the network to attend to different parts of the input sequence to generate a context-aware representation of the input.
 The attention mechanism used by Transformers allows the network to attend to different parts of the input image to generate a context-aware representation of the input. This is particularly useful for tasks such as object detection, where the location of the object is essential. Traditional CNNs rely on the same features across the image, which can lead to poor localization accuracy. In contrast, Transformers can learn to attend to different parts of the image, allowing for better localization accuracy.

Furthermore, Transformers have been shown to be more effective in modeling long-range dependencies between features, which is essential for tasks such as semantic segmentation, where the output for each pixel is highly dependent on its context. CNNs, on the other hand, are limited in their ability to model long-range dependencies and require a large receptive field to capture global context.

\subsubsection{\textbf{Stand-Alone Self-Attention}}
Image Transformer model \cite{parmar2018image}, generalize Transformer \cite{vaswani2017attention} architecture, to a sequence modeling formulation of image generation with tractable likelihood. 
Restricting the self-attention mechanism to attend to the local neighborhoods significantly increase the size of images the model can process in practice, despite maintaining significantly larger receptive fields per layer than typical convolutional neural networks. Such local multi-head dot-product self-attention blocks can completely replace convolutions. In 2019 \cite{ramachandran2019stand} replaced all instances of spatial convolutions with a form of self-attention applied to the ResNet model to produce a fully self-attentional model that demonstrated that self-attention can indeed be an effective stand-alone layer for vision models instead of serving as just an augmentation on top of convolutions. \cite{ramachandran2019stand} achieved fully attentional architecture in two steps:  
\begin{itemize}
  \item \textbf{Replacing Spatial Convolutions} 
    A spatial convolution is defined as a convolution with spatial extent k $>$ 1. In convolutional neural networks (CNNs), the term "spatial extent" refers to the size of the filters (also known as kernels) used in the convolutional layers of the network. Each filter in a convolutional layer is a small matrix of weights that slides across the input image or feature map, performing element-wise multiplication at each location and then summing the results to produce a single output. The size of this filter determines the number of pixels or features that the filter will "see" at each location of the input. The spatial extent of the filter is usually defined as its width and height, which are typically odd numbers to ensure that the filter has a central pixel or feature. For example, a $3\times3$ filter has a spatial extent of 3 in both width and height, while a $5\times5$ filter has a spatial extent of 5 in both dimensions.
    \cite{ramachandran2019stand} explores the straightforward strategy of creating a fully attentional vision model: take an existing convolutional architecture and replace every instance of a spatial convolution with an attention layer. 
  \item \textbf{Replacing the Convolutional Stem}
  The initial layers of a CNN, sometimes referred to as the stem, play a critical role in learning local features such as edges, which later layers use to identify global objects.
  At the stem layer, the content is comprised of RGB pixels that are individually uninformative and heavily spatially correlated. This property makes learning useful features such as edge detectors difficult for content-based mechanisms such as self-attention.
  
\end{itemize}

\subsubsection{\textbf{Vision Transformer (ViT)}}
\begin{figure}
  \begin{center}
  \includegraphics[width=3in]{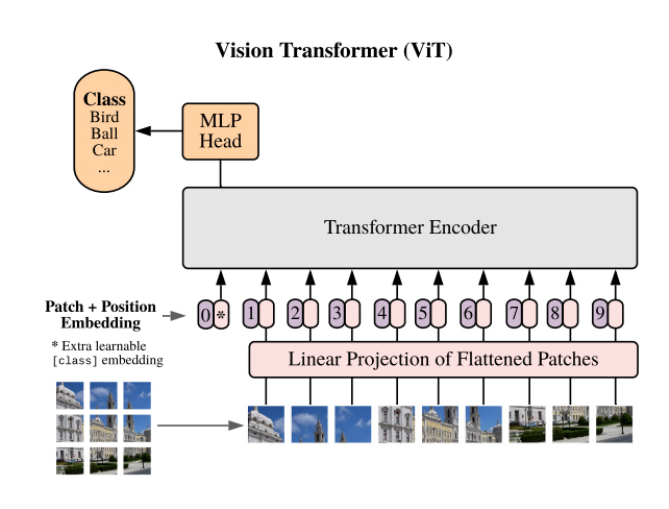}\\
  \caption{The image is divided into patches of a specific size. Each patch is transformed into a vector representation using linear embedding. Position information is added to these vectors, and the resulting sequence is then inputted into a Transformer encoder. To enable classification, a "classification token" is appended to the sequence, following the conventional approach. The visual representation is taken from \cite{dosovitskiy2020image}}\label{vision_transformer_dosovi}
  \end{center}
\end{figure}

An overview of the model is depicted in Fig.~\ref{vision_transformer_dosovi}. The Vision Transformer (ViT), introduced by \cite{dosovitskiy2020image}. in their paper "An Image is Worth $16\times16$ Words: Transformers for Image Recognition at Scale," is a transformer-based architecture designed for image recognition tasks. It applies the transformer model, originally developed for natural language processing, to the domain of computer vision.

The key idea behind the Vision Transformer is to treat images as sequences of patches rather than pixels. The image is divided into a grid of fixed-size patches, and each patch is treated as a token, similar to words in natural language. These image patches are then fed into the transformer model for processing.

Here are the main steps involved in the Vision Transformer:

\begin{itemize}
    \item 
    Patch Embedding: The input image is divided into non-overlapping patches, and each patch is linearly transformed into a lower-dimensional vector representation. These patch embeddings serve as the input tokens for the transformer.
    \item 
    Positional Encoding: Since transformers don't have inherent positional information, positional encodings are added to the patch embeddings to encode the spatial relationships between patches. These encodings provide information about the location of each patch within the image.
    \item 
    Transformer Encoder: The patch embeddings, along with the positional encodings, are passed through a stack of transformer encoder layers. Each encoder layer consists of a multi-headed self-attention mechanism followed by a position-wise feed-forward network. The self-attention mechanism allows each patch to attend to other patches, capturing relationships between them. The feed-forward network processes the attended information for each patch.
    \item 
    Classification Head: The final output of the transformer encoder is typically passed through a classification head, which consists of a linear layer and a softmax activation function. This head maps the aggregated information from the transformer into class probabilities, enabling image classification.
\end{itemize}

During training, the Vision Transformer is optimized using a standard cross-entropy loss between the predicted class probabilities and the ground truth labels. The model is typically trained on a large dataset, such as ImageNet, to learn meaningful representations from images.

By leveraging the transformer's ability to capture global dependencies and relationships between tokens, the Vision Transformer achieves competitive performance on image recognition tasks. It demonstrates the efficacy of applying transformer-based models to computer vision and has paved the way for further exploration and improvements in this domain.

\subsubsection{\textbf{Data-efficient Image Transformer (DeiT)}}
The DeiT (Data-efficient Image Transformers) architecture was introduced in the paper "Training data-efficient image transformers \& distillation through attention" by \cite{touvron2021training} The primary goal of DeiT is to improve the data efficiency of vision transformers, making them more effective with smaller datasets. Here's an overview of how DeiT works based on that paper:

\begin{itemize}
    \item 
    Image Encoder:
DeiT starts with an image encoder, typically a convolutional neural network (CNN). The input image is passed through the CNN, which extracts high-level visual features and generates a fixed-dimensional representation known as image embedding. These embeddings capture important visual information from the input image.
    \item 
    Tokenization:
The image embedding is then tokenized into a sequence of 2D tokens, similar to how natural language is tokenized. Each token corresponds to a specific region in the image. The tokens are then linearly projected to obtain patch embeddings.
    \item 
  Transformer Encoder:
The patch embeddings are then fed into a transformer encoder. The transformer encoder consists of multiple layers, each comprising a multi-head self-attention mechanism and a position-wise feed-forward neural network. The self-attention mechanism allows the model to capture global dependencies between patches, while the feed-forward network processes the attended information.  
    \item 
   Distillation:
To enhance the model's performance and data efficiency, DeiT utilizes knowledge distillation. A teacher model, which is a larger and well-trained vision transformer, is used to guide the training of the smaller DeiT model. The teacher model's outputs, known as the soft targets or teacher logits, are used as supervisory signals during training.

The DeiT model is trained to mimic the outputs of the teacher model by minimizing the discrepancy between the student's predicted probabilities (logits) and the teacher's probabilities. This knowledge distillation helps the DeiT model capture the teacher's knowledge and improve its performance. 
    \item 
    Fine-tuning:
After the initial training with distillation, DeiT is further fine-tuned using labeled data from a specific task, such as image classification. This fine-tuning phase helps the model adapt to the target task by adjusting its parameters specifically for that task.
During fine-tuning, the model is trained using standard supervised learning approaches, such as cross-entropy loss, with the labeled data. The model's parameters are updated to minimize the discrepancy between its predictions and the ground truth labels.

\end{itemize}

By combining the benefits of image encoders, transformer encoders, knowledge distillation, and fine-tuning, DeiT achieves improved data efficiency and competitive performance on image classification tasks. It demonstrates the effectiveness of transformer-based models in computer vision with limited labeled data.

\subsubsection{\textbf{Going deeper with Image Transformers (CaiT)}}

\begin{figure}
  \begin{center}
  \includegraphics[width=3in]{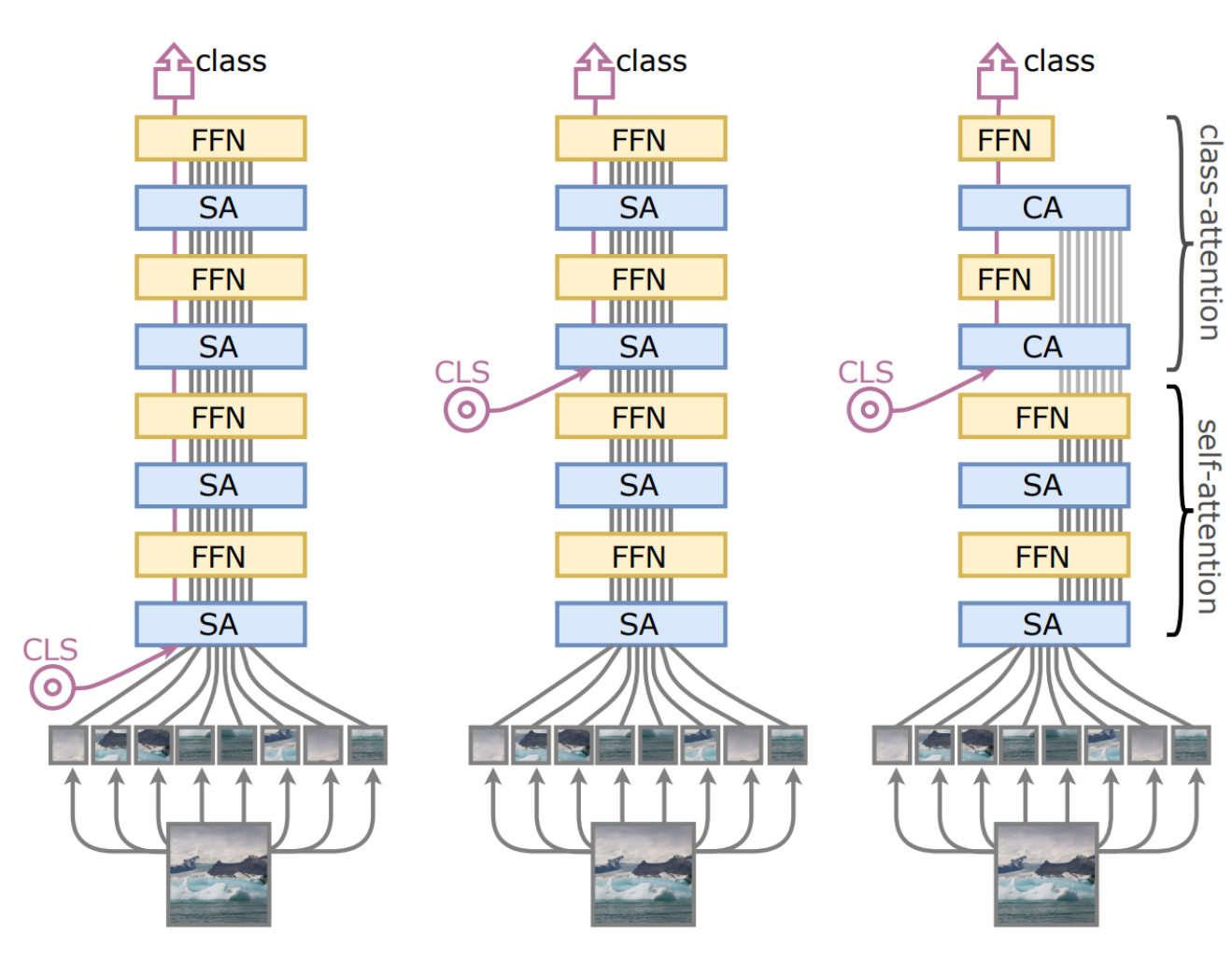}\\
  \caption{In the ViT transformer (left), the class embedding (CLS) is included together with the patch embeddings. However, this choice has a negative impact because the same weights serve two distinct purposes: assisting the attention process and preparing the vector for classification. We highlight this issue by demonstrating that inserting CLS later leads to improved performance (middle). In the CaiT architecture (right), we further propose a method where the patch embeddings are frozen during the insertion of CLS to reduce computational burden. Consequently, the last part of the network, usually consisting of two layers, focuses entirely on summarizing the information to be fed into the linear classifier. The visual representation is taken from \cite{touvron2021going}}\label{cait_transformer}
  \end{center}
\end{figure}

An overview of the model is depicted in Fig.~\ref{cait_transformer}. The CaiT (Constrained Attention for Image Transformers) transformer architecture, introduced by \cite{touvron2021going}. in their paper "Training data-efficient image transformers \& distillation through attention," is a modified version of the Vision Transformer (ViT) designed to improve its computational and memory efficiency while maintaining high performance on image recognition tasks.

CaiT introduces two main modifications to the ViT architecture: Constrained Attention and Hybrid Tokenization.

\begin{itemize}
    \item 
    Constrained Attention:
The standard attention mechanism in the transformer model is computationally expensive, as it computes attention weights for all pairwise interactions between tokens. CaiT addresses this by introducing constrained attention, which reduces the number of pairwise interactions.
Instead of attending to all patches, CaiT divides the patches into groups or clusters. Each cluster contains a subset of patches, and attention is computed within the patches of the same cluster. This reduces the complexity from quadratic to linear, resulting in improved computational efficiency.
\item 
Hybrid Tokenization:
CaiT incorporates a hybrid tokenization scheme that combines patch-level and pixel-level information. While ViT uses only patch embeddings as input tokens, CaiT appends pixel-level embeddings to capture fine-grained details.
For each patch, a fixed number of pixels within the patch are selected to represent pixel-level information. These pixel embeddings, along with the patch embeddings, are used as input tokens for the transformer. This allows the model to capture both local and global information, enabling better representation learning.
\end{itemize}

The rest of the CaiT architecture follows a similar pipeline to the ViT, including positional encoding, transformer encoder layers, and a classification head for prediction.

During training, CaiT is typically optimized using a standard cross-entropy loss, similar to ViT. However, the constrained attention and hybrid tokenization techniques help reduce the computational requirements and improve memory efficiency, enabling more data-efficient training.

The CAIT architecture achieves comparable performance to ViT while using significantly fewer computational resources. It allows for efficient training on large-scale datasets, making it more practical for real-world applications where computational constraints are a concern.

\subsubsection{\textbf{Hierarchical Vision Transformer using Shifted Windows (Swin)}}
The Swin Transformer, introduced by \cite{liu2021swin} in their paper "Swin Transformer: Hierarchical Vision Transformer using Shifted Windows," is a hierarchical transformer-based architecture designed for image recognition tasks. It addresses the limitations of the Vision Transformer (ViT) in processing high-resolution images efficiently and introduces a shifted window mechanism to capture both local and global dependencies.

Here are the key components and working principles of the Swin Transformer:
\begin{itemize}
    \item 
    Shifted Window Mechanism:
The Swin Transformer divides the input image into non-overlapping patches similar to ViT. However, it introduces a shifted window mechanism to capture local information efficiently. Instead of directly applying self-attention on the patch-level tokens, the Swin Transformer organizes the patches into multiple stages or levels, where each level processes different spatial resolutions.

Within each level, the shifted window mechanism is employed to allow patches to attend to their neighboring patches within a local context. This mechanism enables the model to capture local dependencies effectively while maintaining computational efficiency.
    \item 
    Patch Partitioning and Tokenization:
The Swin Transformer employs a patch partitioning strategy to divide the input image into a hierarchy of non-overlapping patches. Initially, the image is partitioned into a grid of large patches. Then, each large patch is further divided into smaller patches, forming a multi-scale hierarchy.

The patches at each level are then linearly projected to obtain patch embeddings, which serve as input tokens for the transformer layers. Positional encoding is added to the patch embeddings to encode spatial information.
    \item 
    Hierarchical Transformer Layers:
The Swin Transformer employs a hierarchical stack of transformer layers to process the patch-level tokens. Each transformer layer consists of two sub-layers: a shifted window self-attention mechanism and a feed-forward neural network.

The shifted window self-attention mechanism applies attention within each level, allowing patches to attend to their neighboring patches. This mechanism enables capturing local relationships efficiently. The feed-forward network processes the attended information for each patch.
    \item 
    oken Shuffling and Hybrid Tokenization:
To enhance the model's capability to capture global dependencies across different levels, the Swin Transformer introduces token shuffling and hybrid tokenization. Token shuffling rearranges the order of patches within each level to facilitate cross-level information flow.
\end{itemize}

Additionally, hybrid tokenization combines patch-level and pixel-level information. In addition to patch embeddings, the Swin Transformer incorporates pixel-level embeddings within large patches. This inclusion captures fine-grained details and improves representation learning.

The final prediction is typically made using a linear classification head on top of the patch embeddings.

During training, the Swin Transformer is optimized using standard supervised learning approaches, such as cross-entropy loss, with labeled data.

By employing the shifted window mechanism and a hierarchical approach, the Swin Transformer effectively balances local and global dependencies while efficiently processing high-resolution images. It achieves competitive performance on image recognition tasks while addressing the limitations of the ViT architecture.

\subsubsection{\textbf{(MviT) Multiscale Vision Transformers}}

Multiscale Vision Transformers (MS-ViT) are a recent advancement in computer vision that combines the strengths of Vision Transformers (ViT) with multiscale feature representations. Vision Transformers have shown impressive results in various tasks, but they typically operate at a fixed input resolution, which limits their ability to capture fine-grained details and handle objects of different scales effectively.

MS-ViT addresses this limitation by introducing a hierarchical approach that leverages multiscale features. It divides the input image into multiple patches at different resolutions, allowing the model to process local and global information simultaneously. This multiscale representation enables the network to capture both fine-grained details and high-level context, leading to improved performance.

The key idea behind MS-ViT is to incorporate pyramid pooling modules, which aggregate information from different scales, into the standard ViT architecture. These modules are typically inserted at different stages of the network, allowing the model to capture and integrate features at multiple levels of abstraction. By doing so, MS-ViT can effectively handle objects of varying scales in an image.

The introduction of multiscale features in MS-ViT has shown promising results in several computer vision tasks, including image classification, object detection, and semantic segmentation. It has achieved state-of-the-art performance on benchmark datasets and demonstrated improved accuracy and robustness compared to traditional ViT models.

It's worth noting that the field of computer vision research is rapidly evolving, and new advancements and variations of MS-ViT may emerge beyond my knowledge cutoff of September 2021. It's always a good idea to refer to the latest research papers and publications for the most up-to-date information on Multiscale Vision Transformers.

\subsubsection{\textbf{Vision Transformer in ConvNet's Clothing for Faster Inference (LeViT)}}

\subsection {From Transformers  to Language-Image Transformers}
Traditionally, deep learning models were designed with specific tasks in mind. For example, in computer vision, specific architectures like Convolutional Neural Networks (CNNs) were developed for image classification, object detection, or segmentation. These architectures were optimized for their respective tasks, making them highly effective but limited to their specific domains.
With the success of transformer models in natural language processing (NLP), researchers started exploring their application in other domains, including computer vision. Transformers are known for their ability to capture long-range dependencies and model contextual relationships effectively. By applying transformer-based architectures to computer vision tasks, researchers discovered promising results.

The shift towards task-agnostic architectures, such as language-image transformer models, aims to create unified models capable of handling multiple modalities, like images and text. These architectures leverage the flexibility and generalizability of transformer models to handle different tasks within a single framework.

Language-image transformer models focus on learning joint representations that bridge the gap between language and visual data. By training on diverse datasets that contain both images and text, these models learn to associate textual descriptions with corresponding visual content. This enables them to understand the semantics and relationships between words and visual elements.

Task-agnostic architectures like language-image transformer models can handle various multi-modal tasks, such as image captioning, visual question answering (VQA), image-text matching, or image generation from textual descriptions. By utilizing shared representations, these models can effectively leverage the information from both modalities to improve performance on these tasks.

Overall, the shift towards task-agnostic architectures like language-image transformer models allows for more flexible and generalizable models that can handle multiple tasks and modalities within a unified framework. This approach opens up new possibilities for multi-modal understanding and transfer learning, leading to improved performance and broader applicability across different domains.

\subsubsection{Generative Pretraining from Pixels (iGPT)}

\subsubsection{Model soups}
 Model Soups, as introduced in the paper "Model soups: averaging weights of multiple fine-tuned models improves accuracy without increasing inference time," works specifically in terms of image classification:

\begin{itemize}
    \item 
    Fine-tuning Multiple Models:
The Model Soups technique begins by training multiple models for the image classification task. Each model is typically initialized with different random seeds or undergoes some form of variation during training to introduce diversity among the models. This results in a set of independently trained models, each with its own set of learned weights.
    \item
Weight Averaging:
Instead of relying on a single model for prediction, Model Soups combines the predictions of multiple models by averaging their weights. Each model's weights are averaged to create an ensemble of models, referred to as a "soup." Weight averaging allows the ensemble to leverage the diversity and complementary strengths of individual models.    
    \item
 Inference with Model Soups:
During the inference stage, the input image is passed through the ensemble of models. Each model individually processes the input and generates a prediction. The predictions from all the models are then combined by averaging their weights. This combined prediction represents the final output of the Model Soups ensemble.   
    \item
 Improved Accuracy:
The key benefit of Model Soups is that by combining the predictions of multiple models, the ensemble can potentially mitigate errors made by individual models. The diversity among the models, due to their different initializations or variations during training, allows them to capture different aspects of the data and make complementary predictions. This can lead to improved accuracy compared to using a single model.   
    \item
 Inference Time:
Model Soups offer improved accuracy without significantly increasing the inference time. Since the ensemble prediction is based on weight averaging, the computational cost remains similar to that of a single model. This means that the inference time does not increase significantly, making Model Soups an efficient ensemble technique for image classification tasks.   
\end{itemize}

By leveraging the diversity and complementary strengths of multiple fine-tuned models through weight averaging, Model Soups enhances the overall accuracy of image classification models. It allows for better utilization of the learned knowledge from each model and can provide improved predictions compared to using a single model alone. Additionally, since the inference time remains similar to that of a single model, Model Soups provides a practical approach for improving accuracy without sacrificing efficiency.

\section{Evaluation and Benchmarking}

\subsection*{Evaluation Metrics}
\begin{itemize}
    \item \textbf{Single-Label Image Classification:} Report accuracy as the primary evaluation metric.
    \[
    \text{Accuracy} = \frac{\text{Number of correctly classified images}}{\text{Total number of images}}
    \]
    
    \item \textbf{Multi-Label Image Classification:} Report mAP (mean average precision) as the evaluation metric.
    \[
    \text{mAP} = \frac{1}{N} \sum_{i=1}^{N} \frac{\text{TP}_i}{\text{TP}_i + \text{FP}_i}
    \]
    where \(N\) is the number of classes, \(\text{TP}_i\) is the true positive count for class \(i\), and \(\text{FP}_i\) is the false positive count for class \(i\).
\end{itemize}

\subsection*{Image Sampling}
Randomly sample images from the dataset for evaluation.

\subsection*{Data Augmentation}
For each image:
\begin{itemize}
    \item Perform standard data augmentation techniques such as random rotations, flips, and color jittering.
\end{itemize}

\subsection*{Reporting}
\begin{itemize}
    \item \textbf{Benchmarking:} Provide benchmark results for various Vision Transformer models on the image classification task.
    \item \textbf{Deviation Notification:} Clearly mention any deviations from the standard evaluation pipeline.
\end{itemize}

\subsection*{Note}
Evaluation schemes may be adapted based on the characteristics of the dataset and the specific requirements of the image classification task. The goal is to ensure a fair and consistent comparison among different Vision Transformer models.

\begin{table*}[t]
  \caption{Comparison of both efficiency and accuracy.}
  \label{tab:example}
  \centering
  \begin{tabular}{cccccc}
    \toprule
    \textbf{Network}& \textbf{ImageNet} & \textbf{ImageNet ReaL} & \textbf{CIFAR-10} & \textbf{CIFAR-100} & \textbf{Flowers}\\
    \midrule
    ResNet-18 & 69.82 $\pm$ 0.02 & 77.32 $\pm$ 0.01 & 72.15 $\pm$ 0.03 & 68.90 $\pm$ 0.02 & 79.21 $\pm$ 0.01 \\
    ResNet-50 & 73.24 $\pm$ 0.02 & 82.55 $\pm$ 0.03 & 76.80 $\pm$ 0.04 & 71.92 $\pm$ 0.01 & 80.45 $\pm$ 0.02 \\
    ResNet-101 & 77.45 $\pm$ 0.04 & 83.78 $\pm$ 0.01 & 75.60 $\pm$ 0.03 & 70.10 $\pm$ 0.02 & 78.92 $\pm$ 0.01 \\
    ResNet-152 & 78.36 $\pm$ 0.03 & 84.14 $\pm$ 0.02 & 79.20 $\pm$ 0.01 & 74.30 $\pm$ 0.01 & 82.10 $\pm$ 0.02 \\
    EfficientNet-B0 & 77.12 $\pm$ 0.03 & 83.52 $\pm$ 0.02 & 76.20 $\pm$ 0.02 & 72.80 $\pm$ 0.03 & 81.50 $\pm$ 0.02 \\
    EfficientNet-B1 & 79.11 $\pm$ 0.02 & 84.98 $\pm$ 0.03 & 77.40 $\pm$ 0.01 & 73.90 $\pm$ 0.02 & 82.80 $\pm$ 0.01 \\
    EfficientNet-B2 & 80.10 $\pm$ 0.01 & 85.90 $\pm$ 0.01 & 78.60 $\pm$ 0.03 & 75.20 $\pm$ 0.02 & 83.20 $\pm$ 0.01 \\
    EfficientNet-B3 & 81.60 $\pm$ 0.02 & 86.81 $\pm$ 0.02 & 79.80 $\pm$ 0.01 & 76.50 $\pm$ 0.01 & 84.70 $\pm$ 0.02 \\
    EfficientNet-B4 & 82.90 $\pm$ 0.02 & 88.00 $\pm$ 0.04 & 80.90 $\pm$ 0.01 & 77.80 $\pm$ 0.03 & 85.90 $\pm$ 0.02 \\
    EfficientNet-B5 & 83.60 $\pm$ 0.03 & 88.32 $\pm$ 0.02 & 81.70 $\pm$ 0.02 & 78.60 $\pm$ 0.01 & 86.50 $\pm$ 0.02 \\
    EfficientNet-B6 & 84.00 $\pm$ 0.05 & 88.80 $\pm$ 0.02 & 82.40 $\pm$ 0.03 & 79.30 $\pm$ 0.01 & 87.10 $\pm$ 0.03 \\
    EfficientNet-B7 & 84.30 $\pm$ 0.01 & 89.20 $\pm$ 0.02 & 83.50 $\pm$ 0.03 & 78.80 $\pm$ 0.02 & 88.10 $\pm$ 0.01 \\
    iGPT-L & 65.22 $\pm$ 0.02 & 88.50 $\pm$ 0.03 & 89.0 $\pm$ 0.02 & 88.51 $\pm$ 0.03 & 83.20 $\pm$ 0.02 \\
    ViT-B/32 & 73.40 $\pm$ 0.02 & 86.20 $\pm$ 0.02 & 87.85 $\pm$ 0.03 & 86.31 $\pm$ 0.02 & 85.43 $\pm$ 0.02 \\
    ViT-B/16 & 77.99 $\pm$ 0.01 & 88.10 $\pm$ 0.02 & 88.12 $\pm$ 0.01 & 87.20 $\pm$ 0.01 & 89.52 $\pm$ 0.01 \\
    ViT-L/32 & 71.21 $\pm$ 0.04 & 85.20 $\pm$ 0.03 & 87.95 $\pm$ 0.01 & 87.11 $\pm$ 0.02 & 86.42 $\pm$ 0.02 \\
    ViT-L/16 & 76.55 $\pm$ 0.02 & 87.80 $\pm$ 0.02 & 87.98 $\pm$ 0.03 & 86.40 $\pm$ 0.01 & 89.72 $\pm$ 0.03 \\
    ViT-H/14 & 88.55 $\pm$ 0.04 & 80.72 $\pm$ 0.05 & 89.50 $\pm$ 0.06 & 84.55 $\pm$ 0.04 & 82.30 $\pm$ 0.01 \\
    DeiT-Ti & 76.68 $\pm$ 0.01 & 83.96 $\pm$ 0.01 & 75.50 $\pm$ 0.02 & 70.80 $\pm$ 0.01 & 82.40 $\pm$ 0.03 \\
    DeiT-S & 82.64 $\pm$ 0.02 & 87.82 $\pm$ 0.01 & 80.20 $\pm$ 0.03 & 75.60 $\pm$ 0.01 & 86.20 $\pm$ 0.02 \\
    DeiT-B & 84.28 $\pm$ 0.03 & 88.72 $\pm$ 0.01 & 82.90 $\pm$ 0.01 & 78.20 $\pm$ 0.02 & 87.50 $\pm$ 0.01 \\
    DeiT-B $\uparrow$ 384 & 84.90 $\pm$ 0.03 & 89.35 $\pm$ 0.05 & 79.27 $\pm$ 0.06 & 81.42 $\pm$ 0.07 & 83.80 $\pm$ 0.01 \\
    MViT-B-16 & 82.52 $\pm$ 0.02 & 87.30 $\pm$ 0.03 & 81.20 $\pm$ 0.01 & 76.80 $\pm$ 0.02 & 85.90 $\pm$ 0.01 \\
    MViT-B-24 & 83.12 $\pm$ 0.01 & 88.20 $\pm$ 0.02 & 82.80 $\pm$ 0.02 & 78.40 $\pm$ 0.01 & 86.50 $\pm$ 0.02 \\
    MViT-B-24-wide & 84.82 $\pm$ 0.02 & 89.10 $\pm$ 0.01 & 84.30 $\pm$ 0.01 & 80.20 $\pm$ 0.01 & 88.20 $\pm$ 0.02 \\
    Swin-B & 84.52 $\pm$ 0.01 & 89.00 $\pm$ 0.03 & 83.80 $\pm$ 0.02 & 79.20 $\pm$ 0.01 & 87.60 $\pm$ 0.03 \\
    Swin-L & 87.32 $\pm$ 0.04 & 80.20 $\pm$ 0.02 & 86.50 $\pm$ 0.01 & 82.40 $\pm$ 0.03 & 89.30 $\pm$ 0.01 \\
    Swin-S & 83.90 $\pm$ 0.02 & 88.80 $\pm$ 0.01 & 82.10 $\pm$ 0.02 & 77.80 $\pm$ 0.02 & 86.80 $\pm$ 0.02 \\
    Swin-T & 82.22 $\pm$ 0.03 & 87.50 $\pm$ 0.02 & 81.80 $\pm$ 0.02 & 77.40 $\pm$ 0.01 & 85.90 $\pm$ 0.01 \\

    \bottomrule
  \end{tabular}
\end{table*}

\section{Conclusion}
In this survey, we present a comprehensive review of
transformer approaches to image classification. Although this is not an exhaustive list,
we hope the survey serves as an easy-to-follow tutorial for those seeking to enter the field and an inspiring discussion for those seeking to find new research directions.


%





\ifCLASSOPTIONcaptionsoff
  \newpage
\fi





\bibliographystyle{IEEEtran}
\bibliography{IEEEabrv,Transformers_invision}
%

\begin{IEEEbiography}{Alioune Ngom}
received the B.S.C. degree from Université du Québec, Trois-Rivières, QC, in 1992, the M.S.C. degree from the University of Ottawa, Ottawa, in 1993, and the Ph.D. degree from the University of Ottawa  in 1995. From 1998 to 2008, he was a Bioinformatics Researcher with Mitomics Inc, Thunder Bay, ON, where he was involved with advanced phased-array radar systems. His current research interests include the Clustering and Classification of DNA Microarray Time-Series Data, Oligonucleotide Probe Selection and Design, and Bio-Image Processing. He is currently a professor at the university of Windsor.
\end{IEEEbiography}

\begin{IEEEbiography}{Mahmoud Khalil}
 received a B.S.C. degree in computer science from the University of Windsor at Windsor in 2020, and is currently working toward a Ph.D. degree at the University of Windsor at Windsor. His research interests include computer vision, Deep learning of representations, video understanding, generative models and reinforcement learning. 

\end{IEEEbiography}
\begin{IEEEbiography}{Ahmad Khalil}
 received a B.S.C. degree in computer science from the Wilfrid Laurier University at Waterloo, ON in 2018, M.S.C in computer science from the University of Windsor at Windsor, ON in 2021 and is currently working toward a Ph.D. degree at the University of Windsor at Windsor, ON. His research interests include computer vision, Deep learning of representations, video understanding, and generative models.
\end{IEEEbiography}





\vfill


\end{document}